\pdfoutput=1

\documentclass{article}
\usepackage{spconf,amsmath,graphicx}
\usepackage{amssymb}
\usepackage{epsfig}
\usepackage{bm}

\usepackage{xcolor}
\usepackage{soul}
\definecolor{AKN}{rgb}{0.56, 0, 1}

\title{Improving Adversarial Robustness with Hypersphere Embedding and Angular-based Regularizations}
\name{Olukorede Fakorede*  \ \  \ Ashutosh Nirala  \ \ \ Modeste Atsague \ \ \ Jin Tian}
\address{Department of Computer Science, Iowa State University \\ \{fakorede, aknirala, modeste, jtian\}@iastate.edu }
\begin{document}
%
\maketitle
\begin{abstract}
Adversarial training (AT) methods have been found to be effective against adversarial attacks on deep neural networks. Many variants of AT have been proposed to
improve its performance. Pang et al. \cite{pang2020boosting} have
recently shown that incorporating hypersphere embedding (HE) into
the existing AT procedures enhances robustness. We observe that the existing
AT procedures are not designed for the HE framework, and thus fail to adequately
learn the angular discriminative information available in the HE framework. In this paper, we
propose integrating HE into AT with regularization terms
that exploit the rich angular information available in the
HE framework. Specifically, our method, termed angular-AT, adds regularization terms to AT that explicitly enforce
weight-feature compactness and inter-class separation; all
expressed in terms of angular features. Experimental results
show that angular-AT further improves adversarial robustness.

\end{abstract}
\section{Introduction}
\label{sec:intro}
The application of deep neural networks (DNNs) into various
domains has raised skepticism due to the observed vulnerability of DNNs to adversarial examples. Adversarial examples
are produced when small, imperceptible but well-crafted perturbations are added to, e.g., natural images leading to a wrong
prediction by the network. This observed flaw in DNNs has
opened an active research area aimed at improving the robustness of DNNs against these malicious attacks.
Of the many methods proposed to improve the robustness
of DNNs against adversarial attacks, adversarial training (AT)
 \cite{goodfellow2014explaining,madry2018towards}, which requires the introduction of adversarial examples in the training of robust
models, has been found effective. The success achieved by AT has led to an array of
AT variants, e.g. \cite{kannan2018adversarial,wang2019improving,zhang2019theoretically}. Additionally, efforts \cite{carmon2019unlabeled,pang2020boosting} have been made to improve the performance of AT. One idea for improving
the performance of AT is augmenting AT with hypersphere
embedding (HE) \cite{pang2020boosting}. HE involves enforcing discriminative
constraints on a hypersphere manifold. This is done by normalizing the linear layer's weight and the penultimate layer's features, and using an additive angular
penalty.

While the traditional softmax 
cross-entropy loss has the disadvantage
of not explicitly encouraging discriminative learning of features \cite{liu2016large,wang2018cosface}, various angular softmax-cross-entropy loss
functions that incorporate HE have been proposed to address
this limitation. Notable implementations in deep learning
literature that utilize HE include CosFace\cite{wang2018cosface}, ArcFace \cite{deng2019arcface},
and SphereFace \cite{liu2017sphereface}, among others.

A major feature of HE is that it encourages learning angularly discriminative features \cite{wang2018cosface,deng2019arcface,liu2017sphereface}. 
While Pang et al. \cite{pang2020boosting} have demonstrated that incorporating HE improves the performance of
the AT methods by directly applying the HE described in CosFace \cite{wang2018cosface} into existing adversarial training methods, e.g., standard AT \cite{madry2018towards}, ALP \cite{kannan2018adversarial}, and TRADES\cite{zhang2019theoretically}, we observe that these existing AT variants were not originally designed for the
HE framework and were not taking advantage of the abundant
angular information.

In this paper, we propose a new training objective that
integrates HE into AT and exploits the rich angular information in the HE framework. 
Our work is the first to address adversarial robustness exclusively using angular information. Our proposed Angular-AT objective consists of
an angular softmax-cross-entropy loss plus two regularization
terms that: (1) encourage weight-feature compactness by
explicitly minimizing the angle between an adversarial feature
vector and the weight vector corresponding to the true class,
(2) encourage inter-class separation by maximizing
the angles among class weight vectors. Lastly, we perform extensive experimental studies to show the effectiveness
of the proposed method.

\section{Related Works}

\subsection{Hypersphere Embedding}
The standard softmax cross-entropy loss has been argued to
lack sufficient feature discriminative power, especially when
deployed in deep face recognition models \cite{liu2017sphereface}. To address
this limitation, variants of HE, which improve intra-class compactness and inter-class variance, were introduced \cite{wang2018cosface,deng2019arcface,liu2016large}. From a geometrical perspective, HE imposes discriminative constraints on a hypersphere manifold and improves the
learning of angular discriminative features. Popular HE
implementations include CosFace\cite{wang2018cosface}, ArcFace \cite{deng2019arcface},
and SphereFace \cite{liu2017sphereface}.

\subsection{Adversarial Robustness}
Many methods have been proposed to defend against adversarial examples \cite{goodfellow2014explaining, madry2018towards, zhang2019theoretically, wang2023anomaly, carlini2022certified,kannan2018adversarial,wang2019improving}, among which 
\textit{AT} \cite{goodfellow2014explaining,madry2018towards} is considered the most effective. 
In their seminal work,
Madry et al. \cite{madry2018towards} formulated adversarial training as a min-max
optimization problem as follows: 
\begin{equation}\label{eq:ATMinMax}
\min_{\bm{\omega}} \mathbb{E}_{(\bf{x},y) \sim \mathcal{D}} \left[ \max_{B_{\epsilon}(\textbf{x})}L_{\bm{\omega}}(\textbf{x}+\delta, y) \right]
\end{equation}
where $L_{\omega}(.)$ is the loss function, $\omega$ are the model parameters, $y$ is the label of $\textbf{x}$, and $\delta$ denotes an adversarial perturbation constrained by the radius $\epsilon$. The inner maximization typically utilizes the Projected Gradient Descent (PGD)
attack to craft adversarial examples. The outer minimization
minimizes the high loss induced by the adversarial examples.

Prominent variants of AT include \cite{zhang2019theoretically,wang2019improving, kannan2018adversarial},
to cite a few. Various ideas such as \cite{carmon2019unlabeled,wu2020adversarial,pang2020boosting} have also been explored to boost the
performance of AT. Pang et al. \cite{pang2020boosting} have incorporated  HE into adversarial training to boost adversarial
robustness.

Our work follows  \cite{pang2020boosting} in integrating HE into
AT. While  \cite{pang2020boosting} directly introduces HE into existing AT variants
such as PGD-AT \cite{madry2018towards}, TRADES \cite{zhang2019theoretically}, and ALP\cite{kannan2018adversarial}, we introduce a novel adversarial training objective that exploits the rich angular
information available on the hypersphere manifold.

\section{Notation and Preliminaries}
\label{sec:preliminaries}
We denote the training set as $\mathcal{D} = \{\textbf{x}_i ,y_i\}_{i=1}^d$ , where $\textbf{x}_{i} \in  \mathcal{X} \subseteq \mathbb{R}^n$ represents a feature vector, $n$ is the dimension of the feature vectors, $y_i \in \{1, \cdots , K\}$ where  $K$ represents the number of labels, and   $d=|\mathcal{D}|.  $ 
 For a feature vector  $\textbf{x} \in \mathbb{R}^n$, \(\|{\textbf{x}}\|_p = (\sum^n_{i=1}|\textbf{x}_{i}|^p)^{\frac{1}{p}}\). We define the \(\epsilon\)–neighborhood of $\textbf{x}$ as \(B_{\epsilon}(\textbf{x}) : \{\textbf{x}' \in \mathcal{X}: \|{\textbf{x}}' - \textbf{x}\|_p \leq \epsilon \}\). 
Let $f_{\omega}: \mathbb{R}^n \xrightarrow{} \mathbb{R}^K$ denote a deep learning classifier with model parameters $\omega$ that produce the output
  \begin{equation}\label{eq:softmax}
          f_{\omega}(\textbf{x}) = \mathbb{S}({\textbf{W}^T}\textbf{z} + b)
  \end{equation}
where $\textbf{z}$ =  $\textbf{z}(\textbf{x};\omega)$ denotes the extracted features in the penultimate layer with the model parameters $\omega$, the matrix \textbf{W}  = \((\textbf{W}_1,...,\textbf{W}_K)\) is the weight matrix, and  $b$ is the bias term in the linear layer.  $\mathbb{S}(.)$: $\mathbb{R}^K \xrightarrow{} \mathbb{R}^K$ is the softmax function.

\subsection{Hypersphere Embedding}
HE typically involves four operations: setting the bias term in eq.~(\ref{eq:softmax}) to zero, weight normalization (WN), feature normalization (FN), and setting  an additive angular margin (AM). In eq.~(\ref{eq:softmax}), when $b = 0$,  $\textbf{W}^T\textbf{z} + b$ becomes $\textbf{W}^T\textbf{z} =  (\textbf{W}^T_1\textbf{z},...,\textbf{W}^T_K\textbf{z})$. The inner product ${\textbf{W}_k^T}\textbf{z} = \|\textbf{W}_k\|\|\textbf{z}\|\cos{\bm{\theta}_k}$, where $\bm{\theta}_k$ is the angle between $\textbf{z}$ and $\textbf{W}_k$. The WN and FN operations are computed as follows:
 \begin{equation}\label{eq:normal}
  \textrm{WN:}     \hspace{0.1cm}  \widetilde{\textbf{W}_k} = \frac{\textbf{W}_k}{\|\textbf{W}_k\|}, \hspace{0.2cm}  \textrm{FN:}\hspace{0.1cm} \widetilde{\textbf{z}} = \frac{\textbf{z}}{\|\textbf{z}\|}.
  \end{equation}
 After applying the FN and WN operations, we have that 
 ${\widetilde{\textbf{W}}^T}\widetilde{\textbf{z}} =  (\widetilde{\textbf{W}}^T_1\widetilde{\textbf{z}},...,\widetilde{\textbf{W}}_K^T\widetilde{\textbf{z}})$ = $(\cos{\bm{\theta}}_1,...,\cos{\bm{\theta}}_K)$. $\cos{{\bm{\theta}}_k}$ represents the  logit for the class $k$ and ${\bm{\theta}}_k$ is the angle between feature $\textbf{z}$ and class weight ${\textbf{W}}_k$. Let $\cos{\bm{\theta}}$ denote $(\cos{\bm{\theta}}_1,...,\cos{\bm{\theta}}_K)$. For  a neural network with hypersphere embedding,  we  rewrite eq.~(\ref{eq:softmax}) as:
 \begin{equation}\label{eq:hypersoftmax}
          \widetilde{f}_{\omega}(\textbf{x}) = \mathbb{S}({\widetilde{\textbf{W}}^T}\widetilde{\textbf{z}}) = \mathbb{S}(\cos{\bm{\theta}}),
 \end{equation}
 where $\widetilde{f}_{\omega}$(\textbf{x}) is the output of the neural network with hypersphere embedding, which we shall refer to as HE-DNN from now on. Wang et al \cite{wang2018cosface} proposed to train HE-DNN with the following cross-entropy loss with angular margin:
 \begin{equation}\label{cross_entropy_embed} 
     L_{CE}(\widetilde{f}_{\omega}(\textbf{x}), y) = - \textbf{1}_y^T log\ \mathbb{S}(s \cdot (\cos{\bm{\theta}} - m \cdot\textbf{1}_y)), 
 \end{equation}
 where the hyperparameter $s > 0$ is a scaling parameter for improving numerical stability during training \cite{wang2017normface,pang2020boosting}, $m$ is the angular margin. 

\section{Proposed Method}

\label{sec:proposed method}

It can be observed from eq.~(\ref{eq:hypersoftmax}) 
that the output logits and
the resulting posterior probabilities of a HE-DNN classifier
depend on the angles between the normalized weight vectors
of the linear layer and the normalized feature vector in the
penultimate layer. Similarly, it can be argued that an adversarial attack crafted on HE-DNN attacks these angles.


Given an example $x$ with label $y$, the goal of an adversarial attack is to craft adversarial example $x'$ from $x$ that fools the classifier to classify $x'$ as $y'$ such that $y'$ $\neq$ $y$. 
Consider a binary HE-DNN classifier with a single output such that the cross-entropy loss aims to maximize ${\widetilde{\textbf{W}}^T}\widetilde{\textbf{z}}= \cos\theta$ on input $x$ with label $y=2$. If $x$ is correctly classified, then $\cos\theta > 0$. However, the adversarial goal of crafting $x'$ becomes making $\cos\theta < 0$, thereby attacking angle $\theta$ between the normalized feature vector $\widetilde{\textbf{z}}$ and the weight vector $\widetilde{\textbf{W}}$.


Given that the angles between the feature vector and weight vectors  contain abundant discriminative information \cite{deng2019arcface,wang2019pr,wang2020dma} and  adversarial attacks attack these angles, we propose a regularization term that directly encourages  the weight-feature compactness, more specifically, by minimizing the angle between adversarial feature vector and the weight vector corresponding to the ground-truth label $y$. In addition, prior works \cite{li2021improving} have argued strong connections between adversarial robustness and inter-class separability. 
We therefore propose an additional angular-based regularization term that improves the inter-class separability.

\subsection{Weight-Feature Compactness}

\begin{figure}[ht]
\includegraphics[width=0.44\textwidth]{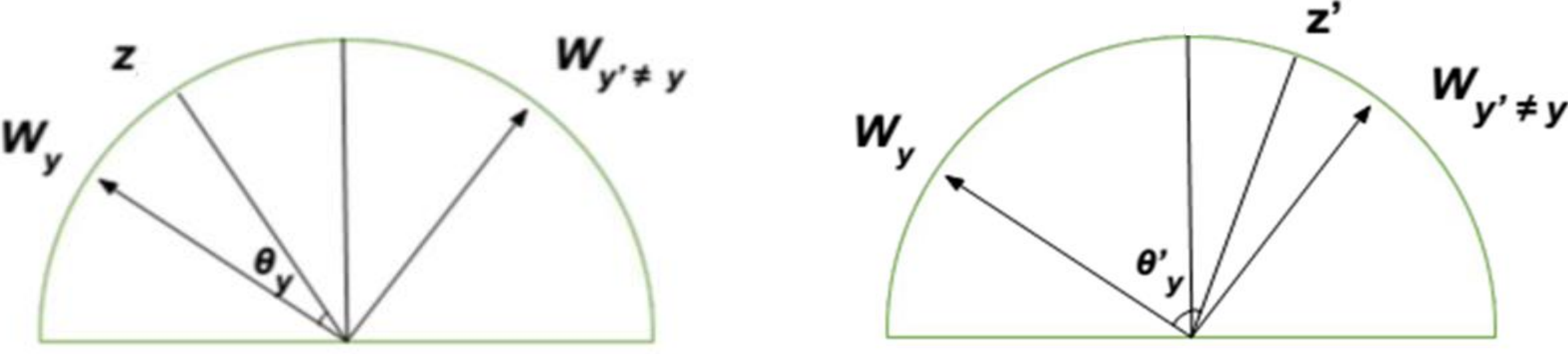}
\caption{The figure on the left shows nearness of a feature vector $\bm{z}$ to the weight vector of the ground truth. The figure on the right depicts how  the resulting feature vector of the corresponding adversarial example, $\bm{z'}$, is pushed towards the weight vector of wrong class $\bm{y'}$ $\neq$ $\bm{y}$, thereby increasing the angle between the feature vector of the adversarial example and the weight vector of the true label.}
\label{fig:geometric-interpretation}
\end{figure}

Generating adversarial examples involves minimizing a model's confidence on an input example w.r.t its true class. Thus, in HE-DNN, the output logit (cosine value) of an input corresponding to its true label is degraded by an adversarial attack. The lower cosine value occasioned by an adversarial attack corresponds to a larger angle between the feature embedding of the adversarial input and the weight vector  of the true label, and consequently a smaller angle between the feature embedding and the weight vector of a wrong class. Hence, there exists a connection between weight-feature   angular compactness and robustness. We provide a geometrical illustration of weight-feature angular compactness in Fig. \ref{fig:geometric-interpretation}. 

To improve robustness, we utilize a regularization term  that encourages  minimization of  the angle between the adversarial feature embedding and the weight corresponding to the ground-truth label $y$. We define the following regularization term to achieve this goal:
\begin{equation}\label{eq:cosarc}
    {l}_{wfc} = [arccos({\widetilde{\textbf{W}_y}^T}\cdot\widetilde{\textbf{z}'})]^2 = (\bm{\theta}'_y)^2 
\end{equation}
where  $\bm{\theta}_y'$ is the angle between the feature embedding of the adversarial example $\textbf{x}'$ and weight vector $\bm{W}_y$ of true class $\bm{y}$. 

\subsection{Inter-Class Separation}
Here we consider the inter-class separation in HE-DNN. 
The weight matrix $\bm{W}$, in the fully connected layer
of a neural network, conceptually represents the various class
centers \cite{wang2018cosface,deng2019arcface}. By
penalizing the angles between these class centers on the hypersphere, we aim to improve angular discriminability and
robustness of HE-DNN to adversarial attacks. We propose to uniformly distribute the class centers around the unit hypersphere. 
To achieve this goal, we utilize the popularly
known Tammes problem \cite{tammes1930origin,lovisolo2001uniform}.

Tammes problem attempts to uniformly arrange n points
on a unit sphere by maximizing the minimum pairwise distance.
Using an idea similar to the Tammes problem, we penalize
minimal angles between class centers by explicitly maximizing the minimum angles between class centers.
Given a linear layer weight matrix $\textbf{W} = (\textbf{W}_1,..., \textbf{W}_K)$,
where $\textbf{W}_k$ corresponds to the weights for class k, we aim
to maximize the angles between each close pair of weight
vectors. We minimize the maximum cosine similarity between pairwise weight vectors by the following regularization term:
\begin{equation}\label{eq:interClassCos}
    {l}_{sep} =  \frac{1}{K}\sum_{i=1}^{K} max_{j,j \neq i}\textbf{C}_{i,j} 
\end{equation}
where $\textbf{C}_{i,j} = \widetilde{\bm{W}}_i^T \cdot \widetilde{\bm{W}}_j$     
represents pairwise similarities, $K$ is the number of classes, and $\widetilde{\bm{W}_i} = \frac{\bm{W}_i}{\|\bm{W}_i\|}$. Globally computing maximum cosine similarity $max_{i,j, i \neq j}\textbf{C}_{i,j}$ 
is inefficient; thus, we compute the mean of each vector’s maximum cosine similarity in eq. (\ref{eq:interClassCos}).


Prior work such as \cite{mettes2019hyperspherical} adopted the Tammes problem in non-adversarial settings to uniformly separate prototypes which are \textit{a priori} positioned on hyperspherical output. In this work, we uniquely apply Tammes-inspired uniform separation to improve the uniform separation of class identities in the linear layer of HE-DNNs.

\subsection{Training Objective}
We combine the regularization terms described in sections 4.1 and 4.2 with the loss function decribed in eq.~(\ref{cross_entropy_embed}) to arrive at the following Angular-AT training objective: 

\begin{equation}
   \min_{\bm{\omega}}\mathbb{E}_{(\textbf{x},y) \sim \mathcal{D}}\Bigl\{ L(\widetilde{f}_{\omega}(\textbf{x}'), y) + {\alpha} l_{wfc} + {\beta} l_{sep}\Bigl\}
\end{equation}
where $\alpha$ and $\beta$ are regularization hyperparameters. $\textbf{x}'$ is an adversarial example crafted by perturbing input example $\textbf{x}$ using  PGD attack. 
$y$ is the true label of $\textbf{x}$.

\section{Experiments}
We experimentally verify the effectiveness of the our method, and compare the
robustness obtained with the state-of-the-art defenses on \textit{CIFAR10/100} and \textit{TinyImagenet}.

\textbf{Baselines.}  We  compare our proposed Angular-AT method against the two best performing defenses that are based on  HE-DNN: PGD-AT-HE and TRADES-HE \cite{pang2020boosting}.  For completeness, we also compare against original defenses PGD-AT \cite{madry2018towards} and TRADES \cite{zhang2019theoretically}, which are based on the standard DNNs.

\textbf{Defense settings.} We utilize ResNet18 (\textit{CIFAR100} and \textit{TinyImagenet}) and Wideresnet-34-10 (\textit{CIFAR10})  \cite{he2016deep} for classification. The models are trained for 100 epochs, using mini-batch gradient
descent with momentum 0.9, batch size 128, weight decay 3.5e-3 (ResNet-18) and 5e-4 (Wideresnet-34-10). The learning rates are set to 0.01 and 0.1 on ResNet-18 and Wideresnet-34-10 respectively. In both cases, the learning rates are decayed by a factor of 10 at 75th, and
then 90th epoch. The adversarial examples used for training are obtained by perturbing each image using PGD attack, setting the perturbation $\epsilon$ = 0.031, the
perturbation step size to 0.007, and the number of iterations to 10.

\textbf{Hyperparameter settings.} For the baselines  PGD-AT-HE and TRADES-HE, $m$ and $s$ are respectively kept at 0.2 and 15.0, as described by the authors \cite{pang2020boosting}. We set $m$ = 0 and the scale $s$ = 15.0. 
 The values of regularization hyperparameters $\alpha$ and $\beta$ are heuristically determined, and the values yielding the best results after multiple experiments were selected. Hence,  $\alpha$ and $\beta$ are set to 0.55, 0.48 respectively.

\textbf{Evaluation settings.} We evaluated our defense   under \textit{White-box attack} threats including PGD-20/500 \cite{madry2018towards}, CW ($L_{inf}$ version of CW loss optimized by PGD20) \cite{carlini2017towards} and Autoattack \cite{croce2020reliable}. The perturbation size is set to $\epsilon$ = 0.031, step size $\eta$ = 0.003. 
In addition, we evaluated the proposed defense on the SPSA\cite{uesato2018adversarial} attack (a strong query-based \textit{Black-box attack}), with the perturbation size of
0.001 (for gradient estimation), sample size of 128, 80 iterations, and learning rate 0.01. Experiments were reproduced four times with different random seeds; the mean and standard deviation are subsequently computed. The results are reported as mean ± std in the tables.  The results, as reported in tables \ref{table:cifar10-whitebox}-\ref{table:tinyimagenet-whitebox}, shows
that our method significantly improves the baselines. 

\textbf{Ablation Studies.} We study the contribution of each regularization to the robustness against PGD, CW, and AA attacks, using CIFAR10 on WRN-34-10. Our observations are reported in table \ref{ablation}. Training HE-DNN using only $L(\widetilde{f}_\omega(x'), y)$ yields good performance on PGD attacks, but weaker performance on stronger attacks like CW and AA attacks. $L(\widetilde{f}_\omega(x'), y)$ + $l_{wfc}$ yield significantly better performance on CW and AA attacks, slightly at the expense of PGD attack robustness. The $l_{sep}$ term improves robustness to PGD attacks, therefore $l_{sep}$ compensates for the drop in PGD robustness caused by the $l_{wfc}$ term.

\begin{table}[!h]
\caption{Robust accuracy (\%) for CIFAR10 on WRN34-10.}
\label{table:cifar10-whitebox}
\setlength{\tabcolsep}{0.8pt}
\vskip -0.10in
\begin{center}
\begin{small}
\begin{sc}
\begin{tabular}{lccccc}

\hline
Defense &  Natural & $PGD^{20}$ &$PGD^{500}$&CW& AA\\
\hline
 Pgd-at  &  86.28 & 55.96&51.83&54.79&51.80\\
 Pgd-at-he  & 86.19  &59.36 &57.49&55.12&51.65 \\
 Trades & 84.62 & 56.48&54.84&54.17&\textbf{53.06}\\
 Trades-he      & 82.98  & 61.34&60.31&54.92&52.11 \\
 \textbf{ANG-AT}    &  \textbf{86.52}\tiny{$\pm$ 0.12} & \textbf{64.01} \tiny{$\pm$ 0.17}&\textbf{63.28}\tiny{$\pm$ 0.10}&\textbf{56.25}\tiny{$\pm$ 0.17}& 52.69\tiny{$\pm$ 0.14} \\
 \hline
\end{tabular}
\end{sc}
\end{small}
\end{center}
\vskip -0.45in
\end{table}

\begin{table}[!h]
\caption{Robust accuracy (\%) for CIFAR100 on RN18.}
\label{table:cifar100-whitebox}
\setlength{\tabcolsep}{0.6pt}
\vskip -0.10in
\begin{center}
\begin{small}
\begin{sc}
\begin{tabular}{lccccc}
\hline
Defense &  Natural  & $PGD^{20}$ &$PGD^{500}$& CW& AA\\
\hline
 Pgd-at  &  56.02 & 27.72&25.35&25.52&23.89\\
 Pgd-at-he  & 55.29  & 30.63 &27.85 &26.78&23.55 \\
 Trades & 53.60 & 28.65 &25.73 &25.48&24.65\\
 Trades-he      & \textbf{56.95}  & 31.40  &28.62 &26.96&24.10 \\
 \textbf{ANG-AT}        & 56.89 \tiny{$\pm$ 0.09}  & \textbf{33.51} \tiny{$\pm$ 0.18}& \textbf{31.92} \tiny{$\pm$ 0.11}&\textbf{28.52} \tiny{$\pm$ 0.12}&\textbf{25.32} \tiny{$\pm$ 0.09} \\
 \hline
\end{tabular}
\end{sc}
\end{small}
\end{center}
\vskip -0.4in
\end{table}

\begin{table}[!h]
\caption{Robust accuracy (\%) for TinyImageNet on RN18.}
\label{table:tinyimagenet-whitebox}
\setlength{\tabcolsep}{5pt}
\vskip 0.10in
\begin{center}
\begin{small}
\begin{sc}
\begin{tabular}{lcccc}

\hline
Defense &  Natural  & $PGD^{20}$ &$CW$&\\
\hline
 Pgd-at  &  49.20 & 22.27&20.54&\\
 Pgd-at-he  & 48.91  & 24.72 &21.37 & \\
 Trades & 48.17 & 21.17 &17.43 &\\
 Trades-he      & 49.05  & 25.91  &20.32 & \\
 \textbf{ANG-AT}        & \textbf{50.21 \tiny{$\pm$ 0.09}}  & \textbf{27.27} \tiny{$\pm$ 0.18}& \textbf{22.29} \tiny{$\pm$ 0.09}&\\
 \hline
\end{tabular}
\end{sc}
\end{small}
\end{center}
\vskip -0.3in
\end{table}

\begin{table}[!h]
\centering
\caption{Ablation study acc\%: WRN-34-10 on CIFAR-10.}
\label{ablation}
\setlength{\tabcolsep}{10pt}
\begin{center}
\begin{small}
\begin{sc}
\resizebox{\columnwidth}{!}{
\begin{tabular}{lccccc}
\hline
Loss terms &  $PGD^{500}$ & CW & AA\\
\hline
  $L(\widetilde{f}_\omega(x'), y)$ &62.34\tiny{$\pm$ 0.22}&53.91\tiny{$\pm$ 0.34}&50.97 \tiny{$\pm$ 0.28}&\\
 $L(\widetilde{f}_\omega(x'), y) + \alpha l_{wfc}$      & 61.52\tiny{$\pm$ 0.17} &\textbf{56.85}\tiny{$\pm$ 0.14}& \textbf{53.10}\tiny{$\pm$ 0.19} & \\
 $L(\widetilde{f}_\omega(x'), y) + \beta l_{sep}$      & \textbf{63.37} \tiny{ $\pm$ 0.08} &55.06\tiny{$\pm$ 0.12} & 51.15 \tiny{$\pm$ 0.19} & \\
 $L(\widetilde{f}_\omega(x'), y) + \alpha l_{wfc} + \beta l_{sep}$      & 63.28\tiny{$\pm$ 0.12} &56.35\tiny{$\pm$ 0.27}& 52.69 \tiny{$\pm$ 0.24}& \\
 \hline
\end{tabular}}
\end{sc}
\end{small}
\end{center}
\vskip -0.3in
\end{table}

\begin{table}[!h]
\centering
\caption{Robust accuracy on SPSA attack: WRN-34-10 on CIFAR-10}
\label{spsa}

\vskip -0.9in
\setlength{\tabcolsep}{10pt}
\begin{center}
\begin{small}
\begin{sc}
\resizebox{0.9\columnwidth}{!}{
\begin{tabular}{lcccc}
\hline
Pgd-he &  Trades-he & \textbf{Ang-AT}&\\
\hline
  64.09\tiny{$\pm$ 0.08} &63.74\tiny{$\pm$ 0.12}&64.32\tiny{$\pm$ 0.11}&\\

 \hline
\end{tabular}}
\end{sc}
\end{small}
\end{center}
\vskip -0.3in
\end{table}

\section{DISCUSSION}
\label{sec:discussion}
In section 4, we argued that adversarial attacks fool the HE-DNN by attacking the angle between features and class weights. Hence, we propose a weight-feature compactness regularization term to force a minimal angle between the feature embedding of an adversarial example and the true class weight. Experimental results show that the $l_{wfc}$ term improves robustness against stronger attacks like CW and AA, at the expense of PGD attack. 

We note in section 4, the significance of encouraging inter-class separation. Input samples corresponding to classes that are semantically  closer to other classes are easily misclassified by adversarial attacks. We propose a regularization term that ensures inter-class separation. The angular margin penalty used in \cite{pang2020boosting} ensures inter-class separation between the true class and the closest incorrect class. In contrast, our approach encourages uniform inter-class separation among all class-centers on the hypersphere. Our experiments show that $l_{sep}$  improves robustness against PGD attacks.

In general, incorporating hypersphere embedding into adversarial training significantly improves the robustness against adversarial attacks, especially PGD.  


\section{CONCLUSION}
\label{sec:conclusion}
In this paper, we improved on an existing work in \cite{pang2020boosting}, which incorporates hypersphere embedding into existing adversarial training methods.  We note that the existing adversarial training methods were not originally designed  for the HE framework and were not taking advantage of the angular information available in the HE framework. Consequently, we intuitively justify  and propose two regularization terms that exploit the rich angular information available in the HE framework to improve adversarial training. Subsequently, we extensively tested our method on various strong attacks. The experimental results show the effectiveness of our method in improving the  state-of-the-art.

\vfill\pagebreak



\bibliographystyle{IEEEbib}
\bibliography{strings,refs}

\end{document}